\documentclass[10pt,twocolumn]{article}
\usepackage{graphicx} 
\usepackage[
  backend=biber,
  style=authoryear,
  natbib=true,
  giveninits=true,
  uniquename=init,
  maxbibnames=3,
  minbibnames=3,
  maxcitenames=2,
  mincitenames=1,
  uniquelist=false,
  dashed=false,
  doi=false,
  isbn=false,
  url=false
]{biblatex}

\usepackage{booktabs}
\usepackage[
  letterpaper,
  left=0.77in,
  top=0.9314in,
  textwidth=6.75in,
  textheight=9in
]{geometry}

\newcommand{\rel}[1]{\,{\footnotesize$(\times#1)$}}

\title{Catch It Where It Happens: Real-Time Error Correction During LLM Reasoning}
\author{Rotem Elimelech, Michael Shalyt, Ido Kaminer \\ Technion -- Israel Institute of Technology\\ Haifa 3200003, Israel
}
\date{November 2026}

\begin{document}

\maketitle

\section{Abstract}
Large language models (LLMs) are increasingly trusted with mathematical, scientific and engineering work, yet their mistakes are often confidently written and hard to spot.
A single wrong step tends to propagate, making the model waste inference tokens following a false premise.
Existing safeguards either check only the final answer and restart from scratch, or rely on a verifier LLM that competes for GPU capacity and slows down response time. 
We propose online, deterministic verification: many kinds of reasoning steps can be validated by code, so a step can be verified in real time, as it is written, without any model call or GPU use.
Whenever a check fails, we point out the mistake and let the model continue from the same spot. 
We demonstrate this approach on symbolic mathematics, where derivations already contain explicit expressions that a computer algebra system can check in milliseconds.
Tested on the ASyMOB benchmark and three open-weights models, our online validation improves accuracy by 1.4–5.6 percentage points, corresponding to a 1.8–14.9\% net reduction in incorrect answers, while increasing token usage by only 6–7\%. 
Because the verifier is independent of the model, the approach complements progress in LLMs rather than competing with it: every new model can benefit immediately, and the benefit grows as deterministic checkers extend to more kinds of reasoning steps.

\section{Introduction}

\begin{figure*}[t]
    \centering
    \includegraphics[width=\textwidth]{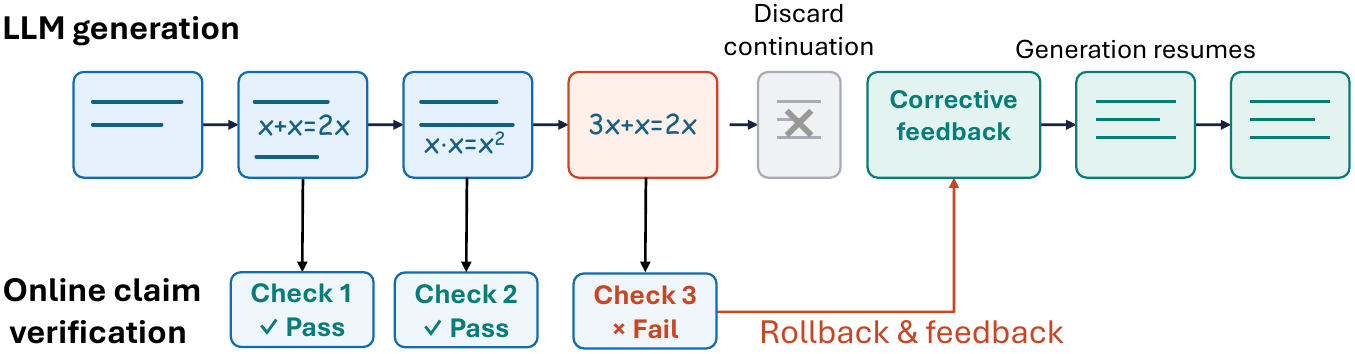}
    \caption{\textbf{Real-time error detection and correction during LLM reasoning.} Mathematical claims are extracted and checked asynchronously on CPU while the LLM continues generating on GPU. Passed checks leave generation uninterrupted. When a check fails, the continuation after the offending step is discarded, corrective feedback is appended, and generation resumes. Verification requires no additional model call or GPU computation.}
    \label{fig:method-overview}
\end{figure*}

As LLMs are being used for increasingly complex and broad tasks in science and engineering, the need for model reliability constantly increases. 
This need has motivated substantial progress in verification-based approaches, both during training and at inference time. 
In formal mathematics, DeepSeek-Prover and AlphaProof use formal verification to train and guide proof generation \citep{xin2024deepseekprover,hubert2025alphaproof}. 
For natural-language reasoning, process supervision trains verifiers to assess intermediate steps \citep{uesato2022process, lightman2024verify, wang2024mathshepherd}, while reinforcement learning with verifiable rewards (RLVR) uses automatically checked outcomes to improve model performance \citep{deepseek2025r1}. 
However, improvements during training do not ensure that each new response is correct, motivating complementary verification during inference.
Studies of intrinsic self-correction have found that prompting models to reconsider their reasoning without external feedback can fail to improve accuracy and can even degrade it \citep{huang2024selfcorrect, kamoi2024selfcorrect}.

Post-generation verification evaluates candidate solutions after they have been produced. Learned verifiers can score completed mathematical solutions and select the most promising candidate \citep{cobbe2021verifiers}. The LLM-Modulo framework couples candidate generation with external verification and feedback-driven revision \citep{kambhampati2024llmmodulo}, while MATH-VF uses an LLM-based formalizer and a tool-integrated critic to check individual statements in a generated mathematical solution and provide feedback for regeneration \citep{zhou2025mathvf}. 
Such approaches can identify errors and improve final answers \citep{gou2024critic, madaan2023selfrefine, shinn2023reflexion, weng2023selfverification}, but checking a completed solution comes after the model has already generated the reasoning that follows an erroneous step. 
Producing multiple candidates or regenerating rejected solutions adds inference cost and can repeat substantial parts of the reasoning \citep{wang2023selfconsistency, snell2025scaling}.

Online verification addresses this timing problem by checking intermediate outputs during generation. Streaming-VR uses an external LLM to verify and refine newly generated text concurrently with the main generation \citep{ko2025streamingvr}. InterWhen extracts intermediate states from an ongoing reasoning trace, checks them asynchronously, and uses rollback and corrective feedback when a violation is detected \citep{bhat2026interwhen}. Model-based monitoring offers flexibility in interpreting free-form text, but requires additional inference: Streaming-VR invokes a verification model, while InterWhen's general reasoning implementation forks the generating model to extract verifiable states. These calls introduce monitoring latency and computational cost even when executed asynchronously. When they share GPU resources with generation, they also compete for inference capacity. Providing the location of a reasoning error can help models correct
mistakes they struggle to identify themselves \citep{tyen2024errors}.

Direct checking of structured outputs can avoid the need for model-based interpretation \citep{poesia2022synchromesh, geng2023grammar}. PICARD uses incremental parsing to reject inadmissible tokens during SQL generation \citep{scholak2021picard}, Monitor-Guided Decoding uses static analysis to constrain code completions \citep{agrawal2023monitor}, and SymStep checks structured atomic deductions with a symbolic constraint propagator \citep{usmanova2026symstep}. These methods enable checks during generation, but operate on formal output languages or prescribed reasoning steps. Their checks therefore do not directly apply to ordinary reasoning traces that mix prose and mathematical expressions. InterWhen also supports direct parsing when suitable structure is available \citep{bhat2026interwhen}. The practical challenge is to obtain the benefits of inexpensive, online checking within free-form reasoning, without requiring additional model inference or a fixed output protocol.
Tool-integrated reasoning uses model-generated programs or tool calls to carry out computations \citep{gao2023pal, chen2023pot, gou2024tora}, which requires the model to decide when to check.

Here we present a method for inexpensive, real-time verification and correction of mathematical claims within free-form LLM reasoning.
Because explicit equations naturally appear within mathematical derivations, they provide checkable claims without requiring a formal language or prescribed step format. 
Following an analogy to speculative execution with branch prediction (in CPU code flows \citealp{smith1981branch}), the model continues decoding on GPU - effectively speculating that its preceding claims are correct - while lightweight CPU parsing extracts completed equations and verifies them asynchronously via a computer algebra system (Figure~\ref{fig:method-overview}).
Passing checks leave generation uninterrupted. 
When a check fails, the short continuation after the erroneous step is discarded, a brief corrective message is appended, and generation resumes from that point.
The checker supplies neither the solution nor a new strategy, but flags a local issue for the model to repair.

We evaluate our method on the ASyMOB symbolic mathematics benchmark \citep{shalyt2026asymob} across three open-weight models and find that online checking improves accuracy across all three by 1.4--5.6 percentage points, with only a 6--7\% increase in mean tokens per answer (due to the model responding to the comment. The amount of discarded tokens is negligible - see Table~\ref{tab:asymob-main}). 
The improvement is concentrated in answers containing detectable errors: accuracy within flagged answers increases by approximately 6--10 percentage points, while performance on unflagged questions remains effectively identical (Figure~\ref{fig:intervention-gains}). 
Where the checker fires, accuracy increases; where it does not, baseline performance is preserved.

The low cost of detection and verification makes this intervention practical on a wide variety of systems. Each check uses approximately 22--25 ms of CPU time, and rollbacks discard only 1.4--15.6 tokens on average across models. Thus, little generated work is lost before a detected error is addressed, and the additional token use primarily reflects continued reasoning and re-derivation following feedback. Extraction and verification are inexpensive enough to run on CPU alongside generation, without requiring another model or competing for GPU capacity. This makes the approach suitable for resource-constrained deployments, where LLM-based verification can substantially reduce generation throughput, and potentially for RLVR training, where checking intermediate steps in every rollout could otherwise add substantial computational overhead. 
More broadly, the same pattern could support online correction in other domains wherever inexpensive checkers can evaluate statements extracted from the generated text.

\section{Method}

\begin{figure*}[t]
\centering
\includegraphics[width=\textwidth]{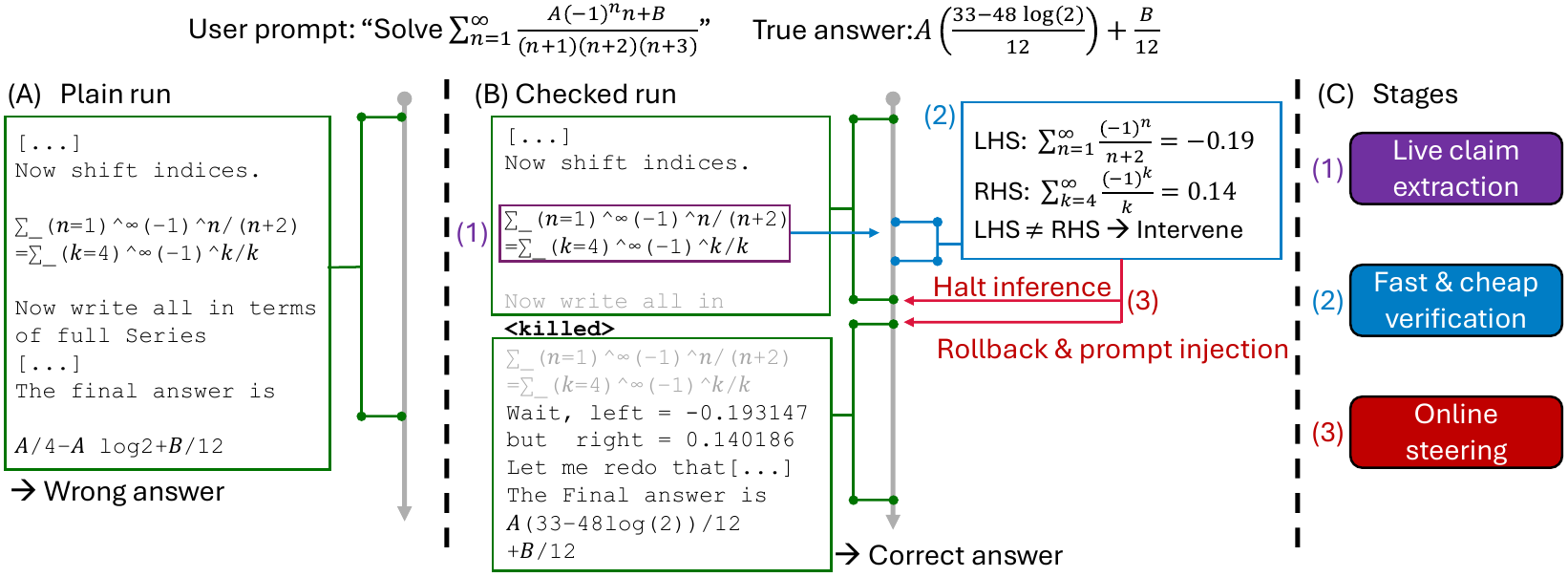}
\caption{\textbf{Live claim extraction, verification, and online steering.}
\textbf{(A)} A plain generation proceeds uninterrupted and produces an incorrect answer.
\textbf{(B)} During checked generation, a completed mathematical claim is extracted from the stream and dispatched for verification while the model continues decoding. A failed check interrupts generation; the continuation after the offending step is discarded, and corrective feedback is appended before generation resumes.
\textbf{(C)} The three stages of the method, corresponding to the numbered annotations in panel~B. Extraction and verification require no additional model calls.}
\label{fig}
\end{figure*}

Our method combines live claim extraction, inexpensive claim verification, and online steering (Figure~\ref{fig}). The model generates a free-form solution while a separate checker evaluates mathematical claims as they become available. Passed checks leave generation uninterrupted. A failed check triggers an interruption, removal of the continuation after the offending step, and insertion of corrective feedback before generation resumes.

The unit of verification is a locally checkable mathematical identity. Each identity is evaluated independently, without reference to the problem statement or the preceding derivation. The method therefore targets \textit{local} errors in mathematical manipulation, such as incorrect signs, expansions, or reindexing. It does not establish whether a step follows from the problem's assumptions or contributes to a valid solution. Misreadings, incorrect strategies, and valid calculations based on a mistaken premise can remain undetected.

\subsection{Live claim extraction}

The model's output arrives in streamed chunks. After each chunk, the extractor scans the text since the last completed mathematical expression and uses regular expressions to identify newly completed expressions. Expressions containing chained equalities, such as $a=b=c$, are split into pairwise claims, $a=b$ and $b=c$. This procedure extracts claims directly from free-form mathematical text without requiring a prescribed step format or an additional model call.

Before verification, the extractor filters out definitions and assignments (identified heuristically as equations with a bare symbol on one side, such as $x=\theta/A$), trivially short or contentless claims, and claims whose two sides contain different sets of free symbols. The latter filter excludes many equations that specify a condition to be solved rather than assert an identity, such as $x^2+2x+1=0$. These filters are heuristics for selecting checkable identities; they can also exclude valid verification opportunities. Previously judged claims are recorded in trace-level state and are not checked again when encountered in subsequent scans or repeated text.

\subsection{Claim verification}

Each extracted claim is passed to a computer algebra system (SymPy \citep{meurer2017sympy}) for numerical evaluation. For expressions without free symbols, the checker compares the numerical values of the two sides. For expressions with free symbols, it evaluates both sides at sampled substitutions and compares the resulting values within a numerical tolerance, following the same numerical comparison procedure used by ASyMOB to grade answers \citep{shalyt2026asymob}. This avoids relying exclusively on symbolic simplification, which may be expensive or unsuccessful for some expressions.

The checker returns one of four outcomes: \textit{holds}, \textit{fails}, \textit{not evaluable}, or \textit{timeout}. Here, \textit{holds} denotes numerical agreement under the performed checks, rather than a formal proof of the identity. Only a \textit{fails} verdict triggers intervention. Expressions that cannot be evaluated or whose check exceeds the time limit leave generation uninterrupted. A failed check can also provide the disagreeing numerical values and, when applicable, the substitution used to obtain them.

This makes checking fast and inexpensive, allowing errors to be detected shortly after they are generated. Verification runs asynchronously on CPU while the model continues generating on GPU, without requiring additional inference or competing for GPU capacity. The checker is independent of the generating model, avoiding reliance on the model's ability to recognize its own mistakes.

\subsection{Online steering}

When a check fails, the controller interrupts generation and retains the text through the end of the offending step. Any text generated after that boundary is discarded. The erroneous step itself remains in the retained prefix so that the appended feedback can refer directly to it.

The controller then appends a short correction phrased as a continuation of the model's own reasoning. Appending text to extend reasoning has also been used in test-time scaling \citep{muennighoff2025s1}. For example:
\begin{quote}
Wait, the line above is not correct: the left-hand side is $-0.193147$, but the right-hand side is $0.140186$. Let me redo that step.
\end{quote}
Generation resumes from this augmented prefix. The feedback identifies the failed claim and its inconsistency without supplying the correct derivation or final answer. The model must determine how to repair the step and continue the solution.

Interventions are bounded by limits of three rollbacks per question. Once the applicable limit is reached, further failures at that step or within that trace do not trigger additional corrections.

\section{Results}

We evaluated model performance on 2,700 easy-to-intermediate problems from ASyMOB \citep{shalyt2026asymob}, using three open-weight models: Qwen3-30B-A3B-Instruct-2507, Qwen3-4B-Instruct-2507 \citep{qwen3}, and Gemma-4-12B-it \citep{gemmateam2026gemma4}. Each problem was evaluated five times per model under two conditions: \textit{plain} generation, where the model runs uninterrupted, and \textit{checked} generation with real-time validation and correction. As shown in Table~\ref{tab:asymob-main}, online checking improved mean accuracy across all three models: from 62.4\% to 68.0\% for Qwen3-30B, from 23.4\% to 24.8\% for Qwen3-4B, and from 45.2\% to 48.2\% for Gemma-4-12B. These gains of 1.4–5.6 percentage points required only a 6–7\% increase in mean tokens per answer, with no additional model calls for verification and no model training. The checker used approximately 22–25 ms of CPU time per check, and rollbacks discarded only 1.4–15.6 tokens on average across models, indicating that intervention followed detected errors after only a short generated continuation. Discarded tokens therefore contributed negligibly to the token budget; the additional tokens primarily reflect the model's continued reasoning and re-derivation following corrective feedback.

\begin{table}[t]
  \centering
  \small
  \setlength{\tabcolsep}{5pt}
  \begin{tabular}{@{}l rr@{}}
    \toprule
    & plain & checked \\
    \midrule
    \multicolumn{3}{@{}l}{\textbf{Qwen3-30B-A3B-Instruct-2507}} \\[1pt]
    answers correct, \%          & $62.4 \pm 0.8$ & $68.0 \pm 0.5$ \rel{1.09} \\
    questions correct all 5      & $796$          & $958$ \rel{1.20} \\
    tokens per answer            & $3{,}320$      & $3{,}520$ \rel{1.06} \\
    tokens discarded at rollbacks& \textemdash    & $2.1 \pm 0.5$ \\
    checker CPU per check, ms    & \textemdash    & $24.6$ \\
    answers intervened on, \%    & \textemdash    & $53.1 \pm 0.8$ \\
    \addlinespace[5pt]
    \multicolumn{3}{@{}l}{\textbf{Qwen3-4B-Instruct-2507}} \\[1pt]
    answers correct, \%          & $23.4 \pm 0.5$ & $24.8 \pm 1.0$ \rel{1.06} \\
    questions correct all 5      & $254$          & $255$ \\
    tokens per answer            & $1{,}753$      & $1{,}877$ \rel{1.07} \\
    tokens discarded at rollbacks& \textemdash    & $15.6 \pm 11.0$ \\
    checker CPU per check, ms    & \textemdash    & $22.1$ \\
    answers intervened on, \%    & \textemdash    & $21.8 \pm 0.9$ \\
    \addlinespace[5pt]
    \multicolumn{3}{@{}l}{\textbf{Gemma-4-12B-it}} \\[1pt]
    answers correct, \%          & $45.2 \pm 0.3$ & $48.2 \pm 0.4$ \rel{1.07} \\
    questions correct all 5      & $428$          & $451$ \rel{1.05} \\
    tokens per answer            & $2{,}579$      & $2{,}748$ \rel{1.07} \\
    tokens discarded at rollbacks& \textemdash    & $1.4 \pm 0.6$ \\
    checker CPU per check, ms    & \textemdash    & $24.7$ \\
    answers intervened on, \%    & \textemdash    & $41.5 \pm 0.4$ \\
    \bottomrule
  \end{tabular}
    \caption{\textbf{Accuracy and inference cost with online verification.} Three open-weight models were evaluated on 2,700 easy-to-intermediate ASyMOB problems, with five repeats per problem under plain and checked generation. Online checking improves accuracy across all three models with a 6–7\% increase in mean tokens per answer. ``Questions correct all 5'' counts problems answered correctly in every repeat; ``answers intervened on'' reports the percentage of answers receiving at least one correction. Parenthetical multipliers give the checked-to-plain ratio. Verification runs entirely on CPU, with little generated text discarded at rollbacks.}  \label{tab:asymob-main}
\end{table}

\begin{figure}[t]
    \centering
    \includegraphics[width=\linewidth]{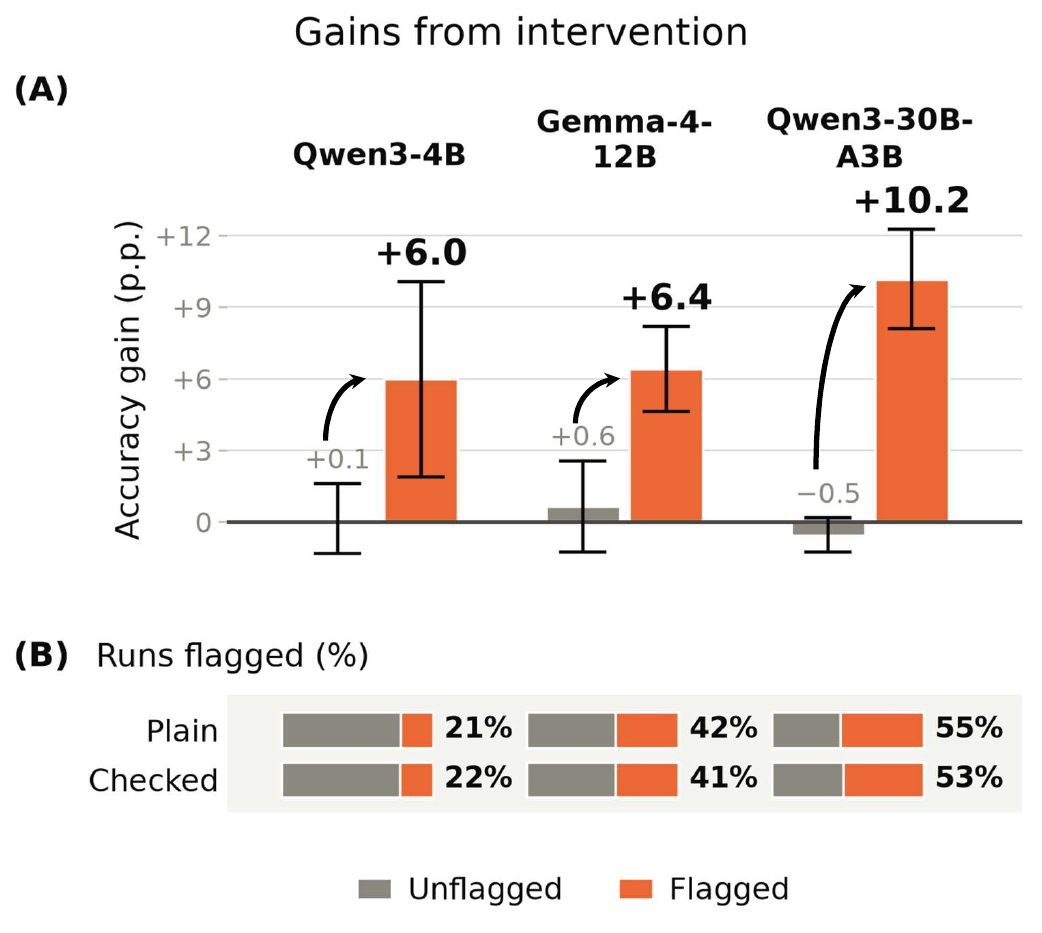}
    \caption{\textbf{Accuracy improves where the checker fires.}
    \textbf{(A)} Checked-minus-plain accuracy differences, in percentage points, for flagged (orange) and unflagged (gray) answers. Accuracy increases by 6.0–10.2 points among flagged answers, while changes among unflagged answers are consistent with zero. Error bars show 95\% confidence intervals over five seeds.
    \textbf{(B)} Fractions of flagged and unflagged answers under each condition; annotations report the flagged percentage. Flags are recorded online during checked generation and identified by replaying the same detection rule offline on completed plain answers.}
    \label{fig:intervention-gains}
\end{figure}

The benefit of online checking depends on two factors: whether the detector fires and whether the resulting correction helps the model reach the right answer. Firing requires the model to express an erroneous step in a form that our extractor and verifier can evaluate. Errors in prose or other uncheckable steps remain outside its reach. To distinguish these factors, we separate answers into flagged and unflagged groups (Figure~\ref{fig:intervention-gains}). In checked generation, a flag triggers an immediate intervention. In plain generation, we replay the same detection rule offline on the completed answer, identifying whether it would have triggered without changing the rollout. This provides a common criterion for comparing answers containing detectable errors under plain and checked generation. The flagged fractions remain similar between the two conditions for each model, differing by at most approximately two percentage points, although the individual answers assigned to each group need not be identical.

Within the flagged group, accuracy increases by 6.0 percentage points for Qwen3-4B, 6.4 for Gemma-4-12B, and 10.2 for Qwen3-30B. These gains, together with the similar flagged fractions, support the interpretation that the method helps models recover from detectable errors rather than simply changing how frequently answers are flagged. Accuracy changes among unflagged answers are close to zero, with all 95\% confidence intervals including zero. This is the expected behavior, as without a flag, no correction is injected and generation follows the ordinary rollout, subject to sampling and execution variability. \textbf{Where the checker fires, accuracy jumps; where it does not, nothing moves within statistical uncertainty}. The overall improvement therefore depends on both the benefit within flagged answers and the fraction of answers reached by the checker: approximately 21–22\% for Qwen3-4B, 41–42\% for Gemma, and 53–55\% for Qwen3-30B.

\begin{figure*}[t]
    \centering
    \includegraphics[width=\textwidth]{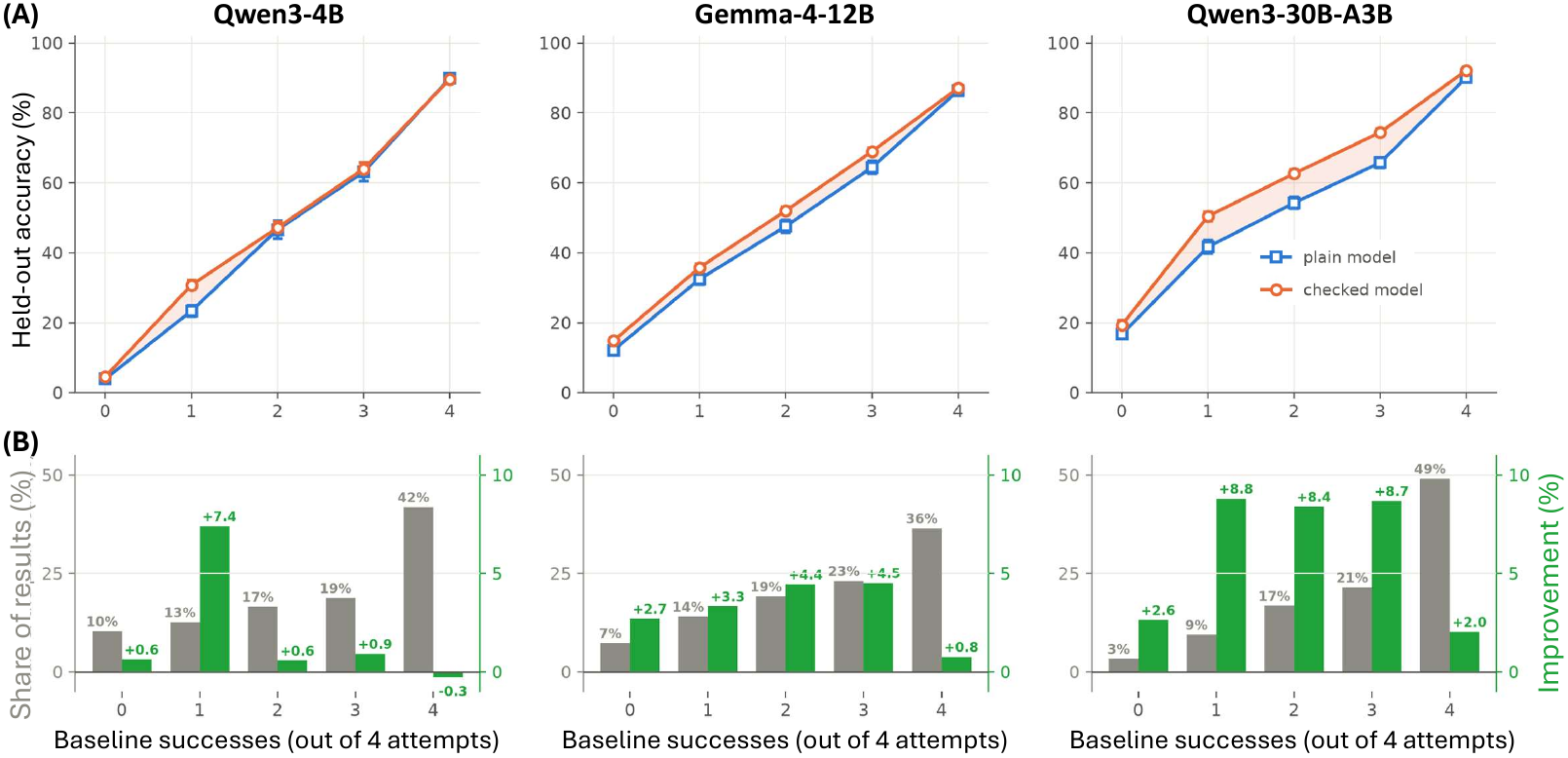}
    \caption{\textbf{Accuracy gains by baseline reliability.}
    Problems are grouped by the number of correct plain-model answers
    across four repeats. Performance is evaluated on the remaining
    repeat and averaged over all five choices of the held-out repeat
    (leave-one-repeat-out evaluation).
    \textbf{Top:} Held-out accuracy for plain (blue) and checked
    (orange) generation. Error bars indicate one bootstrap standard deviation
    across 2,000 resamples of questions, with the full grouping and
    evaluation procedure recomputed for each resample.
    \textbf{Bottom:} Each group's share of all correct plain-model
    answers (gray, left axis) and its accuracy improvement under checking in percentage points (green, right axis).}
    \label{fig:gain-breakdown}
\end{figure*}

Next, we examine how the benefit of online checking depends on the model's baseline reliability, grouping problems by how consistently each plain model solved them across repeated runs (Figure~\ref{fig:gain-breakdown}). For each model, we counted correct answers on four of the five repeats and compared plain and checked accuracy on the remaining repeat, averaging over all five choices of the held-out repeat (leave-one-repeat-out evaluation). Intuitively, this estimates the probability of success on another attempt, given the problem's success rate on the other four attempts, and measures how online checking changes that probability. Keeping the evaluation repeat separate ensures that the outcomes used to define each group are not also used to measure its performance. For Qwen3-30B, the largest gains occurred on problems solved inconsistently by the plain model: the groups with one to three correct answers out of four improved by 8.4–8.8 percentage points and accounted for approximately 82\% of the total net gain. Gemma showed improvements across the zero-to-three-success groups, whereas Qwen3-4B's gain was concentrated in the one-success group, which contributed approximately 67\% of its total net improvement. 

Overall, the gains tend to be smaller at the extremes of baseline reliability and larger on problems the model solves inconsistently. This pattern is consistent with the intended role of our method. We do not train the model and therefore do not add new knowledge or a new problem-solving strategy. Questions beyond the model's capabilities, which fall into the 0/4 group, therefore show limited improvement in performance. Conversely, problems already solved consistently offer less room for improvement and show little net change in accuracy. The main opportunity lies between these extremes, where successful baseline attempts demonstrate that a solution is within the model's reach, but execution remains unreliable. Catching local errors as they occur can help the model realize that existing capability more consistently.

\section{Discussion}
LLMs are used across settings with widely different demands on accuracy, response time, and computational resources. Reliable reasoning matters across these settings, but verification must fit their constraints. When response time is critical, additional model calls can be too costly; when memory and compute are limited, a separate verifier can undermine the advantages of using a smaller model. These constraints motivate verification methods that do not require additional inference.

In mathematics, science, engineering, and code, many intermediate claims can be checked directly through numerical evaluation, symbolic computation, or rule-based tests. Such checks can be inexpensive enough to run alongside generation without invoking another LLM. Our approach uses this opportunity to extract and check mathematical identities on CPU, leaving GPU capacity available for generation. Checks take approximately 22–25 ms each, while the additional token cost primarily comes from the model revising its reasoning after feedback. This separates the resources needed to monitor reasoning from those needed to generate it, making verification practical even when additional model inference would be difficult to accommodate.

Cheap detection is only part of an effective verification system, since the model must also be able to recover from the errors identified. In our experiments, corrective feedback is introduced at inference time, without training the models to respond to this intervention. The checker identifies a local inconsistency, but the model must determine what went wrong, revise the affected reasoning, and continue toward a solution. This suggests an opportunity to co-design models with their verifiers by incorporating verification into training and making effective use of corrective feedback an explicit training objective \citep{kumar2025score}.

Designing systems to support testing and recovery is well established in engineering. Systems often include features specifically intended to make them easier to test, rather than to perform their primary function. In chip design, for example, ``design for testability'' methodology introduces test structures so that manufacturing defects can be detected quickly, reliably, and at low cost \citep{abramovici1994testing}. The analogous goal for language models would be to make reasoning easier to verify by explicitly stating intermediate claims that an inexpensive checker can evaluate. In our experiments, the checker identified a violation in 22–72\% of incorrect baseline answers, depending on the model. Accuracy improved among flagged answers, suggesting that extending verification coverage could allow more answers to benefit from corrective feedback. Training could encourage models to expose checkable intermediate structure even when it is not needed to produce the final answer, increasing the fraction of reasoning accessible to inexpensive verification.

The complementary goal is to improve recovery once an error is detected. In our experiments, 21–51\% of initially incorrect baseline answers with detected violations reached a correct final answer under intervention. In communication systems, checksums provide inexpensive checks for transmission errors, while the surrounding protocol specifies how to respond, for example by requesting retransmission. Similarly, language models could be trained to respond to failed checks by identifying the source of the inconsistency, revising the affected reasoning, and continuing toward a solution. Successfully corrected traces from our method could provide training examples of this recovery process, following the broader approach of learning from successful generated reasoning \citep{zelikman2022star}.

The same approach could extend beyond mathematics wherever generated outputs contain claims that inexpensive tools can evaluate. Examples include syntax and type checks in code \citep{agrawal2023monitor}, dimensional consistency in scientific calculations, arithmetic consistency in financial calculations, and schema validation for tool calls. Each application requires reliable claim extraction and feedback that the model can act on. Mathematics provides a useful first setting because explicit equations already supply many checkable units within free-form reasoning.

Several limitations remain. Local checks cannot establish whether a solution addresses the intended problem, follows a sound strategy, or starts from valid premises. Context-free checking can reject claims that hold under problem-specific assumptions, while numerical sampling, extraction failures, and timeouts limit detection and coverage. Future models trained with verification in mind could help address these limitations by making relevant assumptions explicit and distinguishing checkable claims from steps that require broader interpretation. Such models could combine the strengths of probabilistic reasoning — proposing creative approaches, connecting concepts, and developing informal proofs — with inexpensive verification of the intermediate claims that support their conclusions.

\printbibliography

\end{document}